\documentclass{article}

\usepackage{graphicx}       
\usepackage[utf8]{inputenc}
\usepackage[T1]{fontenc}
\usepackage{hyperref}       
\hypersetup{
	pdftitle={Mathematical artificial data for operator learning},
	pdfauthor={Heng Wu, Benzhuo Lu},
	pdfkeywords={operator learning, PDE, neural network, artificial data},
	pdfsubject={Mathematics, Machine Learning}
}
\usepackage{xcolor}         
\usepackage{amsmath,amssymb,amsfonts}       
\usepackage{booktabs}       
\usepackage{multirow}
\usepackage{pdfpages}
\usepackage{authblk}
\usepackage{caption}
\pdfoutput=1

\title{Mathematical artificial data for operator learning}

\author[1,2]{Heng Wu}
\author[1,2]{Benzhuo Lu\thanks{Corresponding author. E-mail: \texttt{bzlu@lsec.cc.ac.cn}}}

\affil[1]{SKLMS, ICMSEC, NCMIS, Academy of Mathematics and Systems Science, Chinese Academy of Sciences, Beijing 100190, China}
\affil[2]{School of Mathematical Sciences, University of Chinese Academy of Sciences, Beijing 100049, China}

\date{}

\begin{document}
	\maketitle
	
\begin{abstract}
	Machine learning has emerged as a transformative tool for solving differential equations (DEs), yet prevailing methodologies remain constrained by dual limitations: data-driven methods demand costly labeled datasets while model-driven techniques face efficiency-accuracy trade-offs. We present the Mathematical Artificial Data (MAD) framework, a new paradigm that integrates physical laws with data-driven learning to facilitate large-scale operator discovery. By exploiting DEs' intrinsic mathematical structure to generate physics-embedded analytical solutions and associated synthetic data, MAD fundamentally eliminates dependence on experimental or simulated training data. This enables computationally efficient operator learning across multi-parameter systems while maintaining mathematical rigor. Through numerical demonstrations spanning 2D parametric problems where both the boundary values and source term are functions, we showcase MAD’s generalizability and superior efficiency/accuracy across various DE scenarios. This physics-embedded-data-driven framework and its capacity to handle complex parameter spaces gives it the potential to become a universal paradigm for physics-informed machine intelligence in scientific computing.
\end{abstract}

\noindent\textbf{Keywords:}
Mathematical artificial data, Physics-embedded-data-driven, Fundamental solution, Operator learning, Differential equations

\section{Introduction}

Recently, machine learning (ML) has emerged as a revolutionary approach for solving differential equations (DEs), providing novel methodologies that enhance and extend the capabilities of traditional numerical methods \cite{Bi2023, Lu_2021, PhysRevMaterials.2.120301, Schmidt2019, doi:10.2514/1.J058291, Hermann2020, Li2024}. These approaches have shown remarkable success in fields such as weather prediction \cite{Bi2023}, material modeling \cite{PhysRevMaterials.2.120301, Schmidt2019}, fluid dynamics \cite{doi:10.2514/1.J058291}, and quantum chemistry \cite{Hermann2020, Li2024}. ML-based approaches to solving DEs are broadly categorized into data-driven and model-driven methods, depending on their reliance on data or governing equations. Data-driven approaches rely on large, high-quality labeled datasets to train neural networks (NNs) that approximate the mappings between input parameters and solution outputs \cite{Bi2023}. These methods excel in scenarios where extensive datasets are available, such as weather simulation \cite{Bi2023} and turbulence modeling \cite{doi:10.2514/1.J058291}. However, their adoption is limited by the cost and difficulty of acquiring sufficient datasets, particularly for systems requiring high-fidelity solutions or complex geometries. In contrast, model-driven methods, such as Physics-Informed Neural Networks (PINNs) \cite{RAISSI2019686}, incorporate governing equations into the training process, often as constraints in the loss function, ensuring solutions remain physically consistent and interpretable. PINNs and their extensions \cite{Karniadakis_2021} have been applied to diverse areas, including engineering and multiphysics simulations. Despite their promise, model-driven approaches face challenges such as high computational costs, leading to difficulties in handling large-scale systems \cite{Karniadakis_2021, Blechschmidt_2021}.

At the same time, by further leveraging the NNs' flexibility and their capability to capture complex patterns and relationships within data, neural operator learning has been explored for solving parametric DEs. This is a promising direction driven by practical demands—in nearly every real-world application, there is a critical and urgent need for a solver that can rapidly provide solutions to DEs of the same type but with varying parameter functions. Unlike traditional machine learning techniques focusing on finite-dimensional mappings, operator learning targets learning mappings between much larger spaces, in fact infinite-dimensional function spaces, addressing significant limitations of conventional methods. For instance, even minor modifications to boundary conditions or other parameters usually necessitate retraining the entire model from scratch, resulting in inefficiencies and increased computational costs. In contrast, a learned operator can provide approximate solutions to parametric equations (involving boundary conditions and/or other parameters) with complex, high-dimensional, or nonlinear properties. The adaptability and generalization capabilities of learned operators enable transferring solutions across different contexts and scenarios, fostering a more unified and scalable approach to solving equations.

Similarly, operator learning has evolved through both data-driven and model-driven paradigms, as exemplified by notable neural operators such as DeepONet \cite{Lu_2021} (which can be either data-driven or model-driven by embedding PINNs) and the Fourier Neural Operator (FNO) \cite{li2021fourier}. These methods have significantly reduced computational overhead, enabling efficient solutions for high-dimensional systems. However, the challenge of costly data acquisition (whether experimental or computational) has become more pronounced, as it often requires large-scale training datasets, especially for complex systems characterized by high-dimensional parameter spaces \cite{anandkumar2019neural}.

\begin{figure*}[htbp!]
		\centering
		\includegraphics[width=1.0\textwidth]{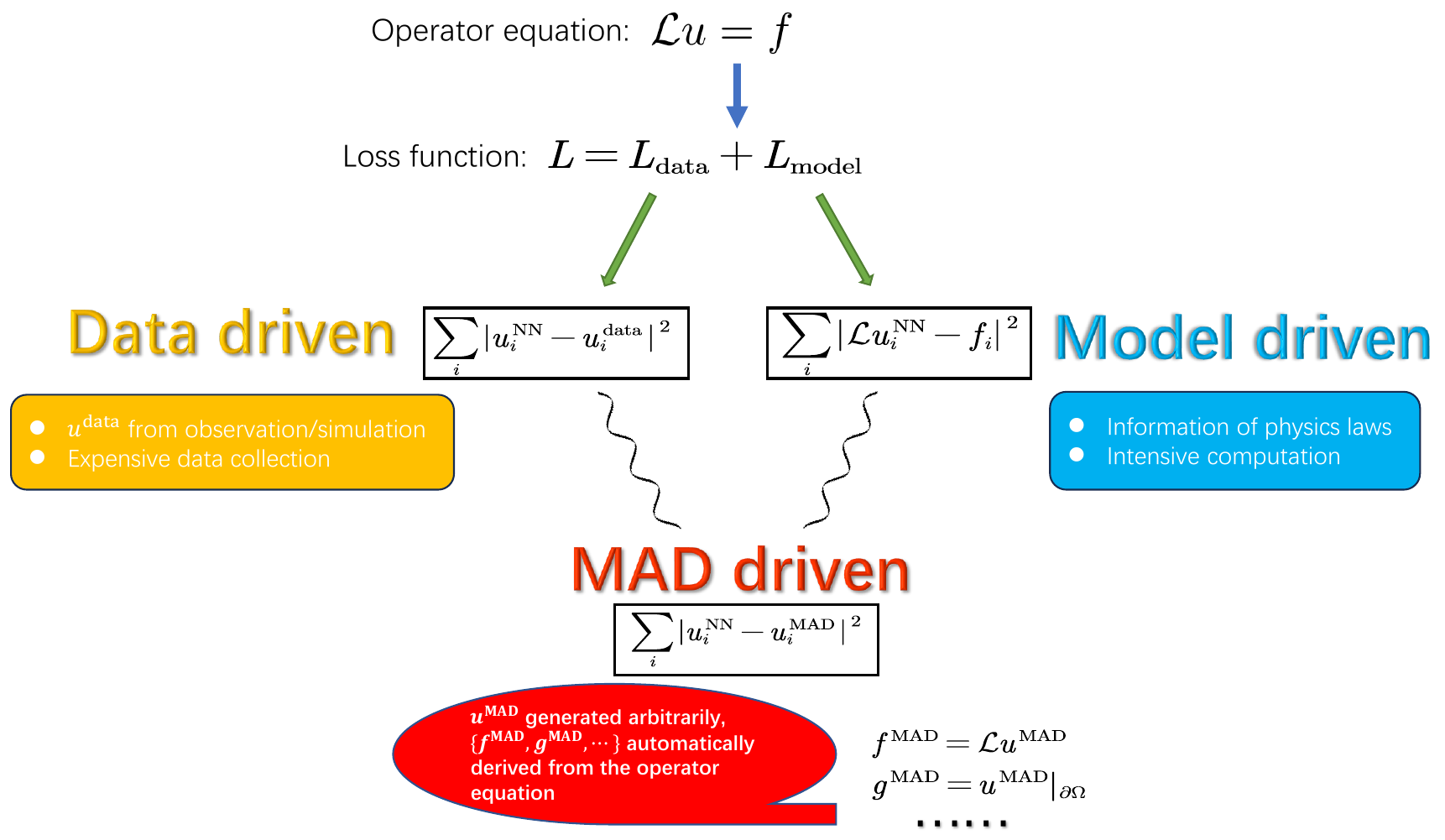}
		\caption{Schematic description of the three paradigms in operator learning, where \( u^{\text{data}} \) denotes the training data from observations or simulations used for supervised learning in the data-driven approach, \( u^{\text{NN}} \) the operator neural network used to approximate the solution \(u\) (operator) of the problem, and \( u^{\text{MAD}} \) the analytical solution generated by MAD method.}
		\label{fig1}
	\end{figure*}

To address these limitations, we propose a new method called Mathematical Artificial Data (MAD), which generates exact analytical solutions rapidly as training data by exploiting the inherent mathematical properties of differential equations. The core idea of MAD is to expand the learning space, decoupling the data generation process from traditional constraints. Our MAD method has some theoretical guarantees, as mentioned in Section~\ref{sec:methods}. Fig.~\ref{fig1} illustrates the characteristics of the three paradigms: data-driven, model-driven, and MAD methods. The primary distinction among these paradigms lies in the formulation of the loss function and the sources of data used. Unlike the other approaches that rely on time-consuming simulation/evaluation data or costly experimental data, MAD delivers rapid, efficient, and cost-effective production of error-free, diverse datasets in a fraction of the time. Moreover, by separating data generation from neural network design, MAD remains inherently flexible and ready to integrate future advances in architectures, optimization algorithms, and other machine-learning innovations.
	
As depicted in Fig.~\ref{fig1}, data-driven methods focus on minimizing the discrepancy between model predictions and observed data, while model-driven methods emphasize ensuring that the model adheres to the governing physical equations. In contrast, the MAD paradigm circumvents the need for external data by generating artificial data directly from operator equations, utilizing mathematical information. This unique approach enables more efficient and accurate data-driven training by mitigating the noise and uncertainty inherent in conventional data sources, offering a more scalable and robust framework for solving complex differential equations.
	
In this paper, we showcase the MAD framework’s core strengths—highly efficient and error-free data generation of its training sets—through extensive numerical experiments on various parametric PDEs. In particular, the fundamental-solution-based MAD1 proposed for source-free equations proves to be even more effective. As a purely data-driven approach, MAD is fully compatible with any modern neural architectures, optimization algorithms, or emerging machine-learning advances. Although here we compare only against classical methods to clearly illustrate the benefits of high-quality physics-embedded artificial data, MAD also has the potential to serve as a powerful data-augmentation or additional-constraint mechanism for both data-driven and model-driven para\-digms—pointing to far-reaching impacts on future differential-equation modeling and operator-learning workflows.

The remainder of this paper is organized as follows. In Section~\ref{sec:methods}, we introduce the Mathematical Artificial Data (MAD) framework, detailing its theoretical foundation, data‐generation procedures, and evaluation metrics. Section~\ref{sec:results} presents numerical results, and Section~\ref{sec:discussion} discusses the implications, limitations, and future directions.

\section{Methodology}
\label{sec:methods}
In this section, we describe the theoretical foundation of the MAD framework,
its data-generation procedures, and the evaluation metrics.

\subsection{Framework for operator learning}

We consider a general partial differential equation (PDE):
\begin{equation}
	\begin{cases}
		\mathcal{L}u = f, & \text{in } \Omega, \\
		u = g, & \text{on } \partial\Omega,
	\end{cases}
	\label{eq00}
\end{equation}
where \(\mathcal{L}\) denotes a differential operator, and \(\Omega\) represents a predefined computational domain, $f$ and $g$ can be general functions. 

Operator learning aims to find the map from inputs (parameters such as initial conditions, boundary conditions, source functions, and/or other parameters appearing in the operator) to solutions of PDEs. To the best of our knowledge, there are only limited research on operator leaning with more than 2 input functions for 2D or higher dimentional PDEs. 

In this work, we demonstrate to use our method to learn the mapping from input function pairs \((f, g)\) to the corresponding solution \(u\) of a general PDE with source (i.e. $\varphi:(f,g)\mapsto u$ )  and the mapping from an input function \(g\) to the solution \(u\) of a source-free PDE (i.e. $\varphi_1:g\mapsto u$ ) .

One approach is to directly use a NN structure like MIONet \cite{doi:10.1137/22M1477751} to represent the mapping from \((f, g)\) to \(u\). Alternatively, as demonstrated in this work, the solution operator for a linear PDE can be decomposed into two components, \(\varphi(f, g) = \varphi_1(g) + \varphi_2(f)\), where \(\varphi_1(g)\) represents the solution to the boundary value problem with \(f = 0\), and \(\varphi_2(f)\) corresponds to the solution of the source term with homogeneous boundary conditions (\(g = 0\)). For equation~\eqref{eq00} as an example, the solution can be expressed as \(u(x) = u_1(x) + u_2(x)\), where \(u_1(x)\) solves \(\mathcal{L} u_1 = 0\) with \(u_1|_{\partial \Omega} = g\), and \(u_2(x)\) solves \(\mathcal{L} u_2 = f\) with \(u_2|_{\partial \Omega} = 0\). This decomposition enables constructing modular NN where boundary condition and source term can be processed separately by corresponding respective modules. Each module can be pre-trained individually (this can greatly reduce the learning space and thus improve efficiency) or combined for training.  Each pretrained operator, such as $\varphi_1(g)$ or $\varphi_2(f)$ can be reused as a component in a appropriate combination settings based on operator learning tasks. This modularity enhances the flexibility and computational efficiency of the operator learning framework, while also being suitable for design to preserve some inherent mathematical properties of each decomposed operator (like the linearity of the operator). The consideration leads to the dual-DeepONet structure construction used in this work (see Fig.~\ref{fig:dual_deeponet} in subsection ~\ref{subsec:setup}).

\subsection{Data generation}
In this subsection, we present the MAD framework’s data-generation procedures—highlighting their efficiency, data-quality guarantees, and seamless compatibility with any modern neural operator architecture—and we also describe the corresponding training-set construction for the PINN-based models to ensure full reproducibility.
\subsubsection{Dataset generated using neural networks for equations with source (MAD0)}
For equations with source, a fully connected neural network with sine activations was used to approximate \(u\). Sine activations were chosen for their universal approximation capabilities \cite{Stinchcombe_1989} and their effectiveness in capturing oscillatory solutions. Specifically, we employed a fully connected network with a [2, 50, 50, 1] architecture, where the first layer has 2 neurons, followed by two hidden layers with 50 neurons each, and the final output layer consists of 1 neuron. All layers used sine activations, enabling the network to sample more complex functions effectively.

It is worth noting that while other activation functions like Sigmoid or Tanh also have universal approximation capabilities, sine activations are particularly effective for generating high-frequency functions. Networks using these activations can more naturally sample functions with oscillatory behaviors. In contrast, networks with other activation functions may still approximate any continuous function, but they might require more careful or specialized initialization to cover a broader function space effectively.

\subsubsection{Fundamental solution-based artificial data (MAD1)}

For linear PDEs with fundamental solutions, the artificial data can be generated based on their fundamental solutions. In this work, we give examples for the Laplace equation and the source-free Helmholtz equation. For the non-zero source cases, the main methodology is still applicable but will not be focused on in this study.

\paragraph{Laplace equation} 
For the Laplace equation, we utilized fundamental solutions constructed using the logarithmic kernel \(\frac{1}{2\pi} \ln \|x - y\|\), which plays a critical role in ensuring comprehensive coverage of the harmonic function space \cite{Bogomolny_1985}. The coefficients of the linear combinations were sampled from a standard normal distribution, introducing variability and ensuring comprehensive coverage of the solution space.

Moreover, for the three-dimensional Laplace equation on \([0,1]^3\), we employed the 3D fundamental solution \(-\frac{1}{4\pi\|x - y\|}\),
and similarly sampled its linear combination coefficients from \(\mathcal{N}(0,1)\).

\paragraph{Source-free Helmholtz equation} 
For the source-free Helmholtz equations (\(f=0\)), solutions were constructed as linear combinations of the fundamental solutions \(J_0\) and \(Y_0\), centered at points outside the computational domain. The zeroth-order Bessel function of the first kind, \(J_0(r)\), and the second kind, \(Y_0(r)\), are defined as:
\begin{equation*}
	J_0(r) = \sum_{m=0}^\infty \frac{(-1)^m}{(m!)^2} \Bigl(\frac{r}{2}\Bigr)^{2m},
\end{equation*}
\begin{equation*}
	Y_0(r) = \frac{2}{\pi} \Biggl[
	J_0(r) \ln\Bigl(\frac{r}{2}\Bigr) \\
	- \sum_{m=1}^\infty \frac{(-1)^m}{(m!)^2} 
	\Bigl(\frac{r}{2}\Bigr)^{2m} \Bigl(\psi(m) + \psi(m+1)\Bigr)
	\Biggr],
\end{equation*}
where \(\psi(m)\) is the digamma function. These functions arise naturally as the real and imaginary parts of the Hankel function of the first kind, \(H_0^{(1)}(r) = J_0(r) + i Y_0(r)\), which serves as the fundamental solution to the 2D Helmholtz equation. While \(H_0^{(1)}\) is widely used in complex-valued formulations, \(J_0\) and \(Y_0\) provide a convenient basis for constructing real-valued solutions.

The coefficients of the linear combinations were sampled from a standard normal distribution, introducing variability and ensuring comprehensive coverage of the solution space. The completeness of such linear combinations in the solution space has been established in the literature, particularly in \(L^2(\partial\Omega)\), where the fundamental solutions form a dense set \cite{10.1093/imamat/30.1.27, Colton_Kress_2019}.

\subsubsection{Another example for constructing harmony functions for Laplace equation (MAD2)}

For the Laplace equation, we use a second sampling strategy based on parameterized trigonometric and hyperbolic functions. Specifically, we utilize linear combinations of \(\bigl(A_i \cos(a_i x) + B_i \sin(a_i x)\bigr)\bigl(C_i \cosh(a_i y) + D_i \sinh(a_i y)\bigr)\), where \(A_i, B_i, C_i, D_i, a_i\) are randomly drawn from the standard normal distribution. This method is referred to as MAD2 in our experiments. This second approach, MAD2, offers a different way of constructing solutions to the Laplace equation using parameterized functions, and serves to highlight the advantages of the fundamental solution-based MAD approach (MAD1).

\subsubsection{Data sampling in PINN-based methods}  
Training datasets for PINN-based models are constructed by independently sampling boundary conditions \(g\) and source terms \(f\), ensuring diversity and representativeness for robust model training.

\paragraph{Boundary conditions sampling}  
Boundary conditions \(g\) are generated using Gaussian Random Field (GRF), characterized by a zero mean and spatially correlated fluctuations. For all boundary shapes, the boundary is first extended into a straight line segment, upon which GRF sampling is performed. After the sampling, a linear function is subtracted to ensure that the boundary condition values at the endpoints match. The resulting boundary conditions are then mapped back to the original boundary geometry. Specifically, \(g\) is defined as:
\[
g \sim \mathcal{G}(0, k_l(x_1, x_2)),
\]
where the covariance kernel \(k_l(x_1, x_2)\) is specified as a radial basis function (RBF):
\[
k_l(x_1, x_2) = \exp\left( -\frac{|x_1 - x_2|^2}{2l^2} \right).
\]
The length-scale parameter \(l = 0.1\) controls the smoothness of boundary condition variations, following the methodologies of Rasmussen \cite{10.7551/mitpress/3206.001.0001} and its applications in DeepONet \cite{Lu_2021}.

\paragraph{Source terms sampling}  
Source terms \(f\) are initially sampled as random noise to ensure variability across the domain. These raw samples are then smoothed using a Gaussian kernel to ensure continuity and smoothness in the training data. The Gaussian smoothing is applied over the entire 2D domain. For each grid point \((x, y)\), the smoothing function is applied in both the \(x\) and \(y\) directions, ensuring that the random field is smoothly varying across the entire space. The smoothed field is then expressed as:
\[
f_{\text{smoothed}}(x) = \frac{1}{\sqrt{2\pi \sigma^2}} \int_{\Omega} f_{\text{raw}}(y) \exp\left(-\frac{|x-y|^2}{2\sigma^2}\right) \, dy,
\]
where \(\sigma = 5\) controls the smoothing scale. This Gaussian smoothing ensures suitable continuity and smoothness for subsequent PINN training, as Gaussian smoothing is a widely used technique in numerical analysis and image processing \cite{10.5555/22881}.

\subsection{Experimental setup}
\label{subsec:setup}
Experiments were conducted on 2D domains, including a square, unit disk, and L-shaped regions. Both MAD and PINN-based methods utilized datasets of 2000 functions, discretized on a \(51 \times 51\) grid for source terms, and 200 boundary points for the square region.

\begin{figure*}[htbp]
	\centering
	\includegraphics[width=1.0\textwidth]{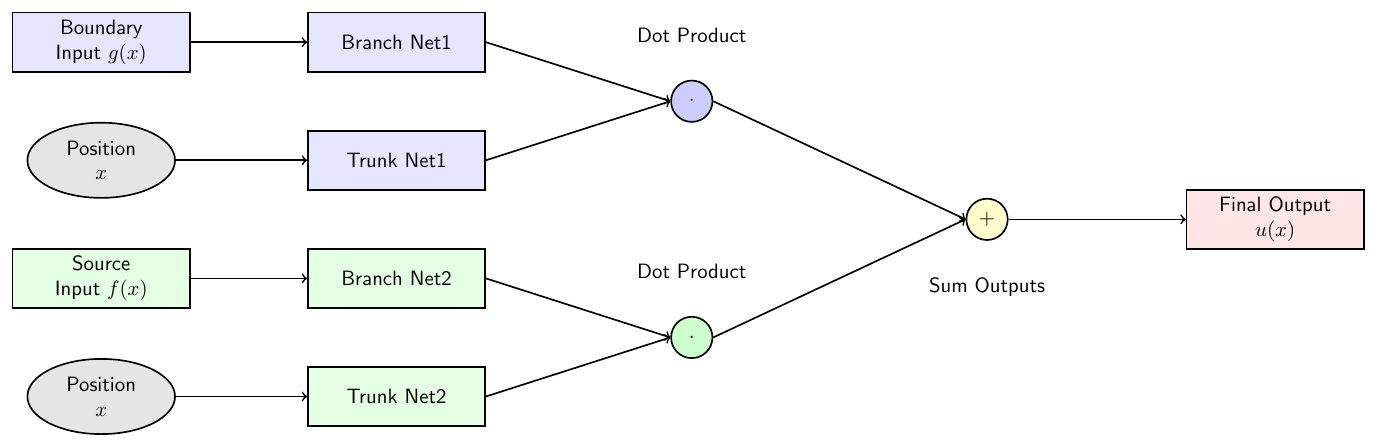}
	\caption{
		\textbf{Architecture of the dual-DeepONet framework.}
		The dual-DeepONet consists of two separate DeepONet components: DeepONet 1 for encoding boundary inputs $g(x)$ and DeepONet 2 for encoding source inputs $f(x)$. Each DeepONet component includes a branch network (encoding inputs) and a trunk network (encoding spatial positions). The outputs of DeepONet 1 and DeepONet 2 are combined through addition to produce the final solution $u(x)$.
	}
	\label{fig:dual_deeponet}
\end{figure*}

The neural network architecture for MAD included:
\begin{itemize}
	\item For source-free equations: An unstacked DeepONet was used with a branch network of size \([200, 1000]\) and a trunk network configured as \([2, 110, \allowbreak 110, 110, 110, 1000]\). This configuration served as the baseline for the following modifications:
	
	\begin{itemize}
		\item Wide configuration: In the wide configuration, the number of neurons in each hidden layer was doubled compared to the baseline configuration, resulting in a branch network of size \([200, 2000]\) and a trunk network with layers of size \([2, 220, 220, 220, 220, 2000]\).
		\item Deep configuration: The deep configuration involved adding more layers to the trunk network. Specifically, the trunk network's hidden layer count was doubled compared to the baseline, leading to a network architecture of \([2, 110, 110, 110, 110, 110, 110, 110, 110, 1000]\), with the same number of neurons per layer as in the baseline configuration.
	\end{itemize}
	
	\item For equations with source terms: As shown in Fig.~\ref{fig:dual_deeponet}, a dual-DeepONet framework was constructed by combining two unstacked DeepONets:
	\begin{itemize}
		\item The first DeepONet, responsible for encoding boundary condition inputs, had a branch network of size \([200, 1000]\) and a trunk network configured as \([2, 90, 90, 90, 90, 1000]\).
		\item The second DeepONet, dedicated to encoding source term inputs, had a branch network of size \([2601, 1000]\) and a trunk network configured as \([2, 110, 110, 110, 110, 1000]\).
	\end{itemize}
\end{itemize}

In all configurations, the branch networks used linear activation functions to preserve the linearity of the operator with respect to the input functions (e.g., source terms \(f\) or boundary conditions \(g\)), ensuring accurate modeling of the linear dependence of the solution on these inputs. On the other hand, the trunk networks employed the \texttt{tanh} activation function to capture the nonlinear behavior of the solution space, leveraging its smooth gradient and bounded output range to approximate complex, nonlinear interactions in the problem domain.

For the 3D Laplace equation over \([0,1]^3\), we employ a DeepONet with:
\begin{itemize}
	\item \textbf{Branch network}: fully connected layers of size \([2402,1000]\), linear activation;
	\item \textbf{Trunk network}: fully connected layers of size \([3,110,110,110,110,1000]\), ReLU activations;
	\item \textbf{Output fusion}: a trainable 1000-dimensional weight vector to combine branch and trunk outputs.
\end{itemize}

For the two MAD1-FNO variants on the 2D Laplace equation over \([0,1]^2\), the input tensor has shape \([C_{\rm in}\times N\times N]\), where \(C_{\rm in}=2\) for MAD1-FNO1 or \(4\) for MAD1-FNO2. The network consists of:
\begin{itemize}
	\item \textbf{Input projection:} 1×1 convolution mapping \(C_{\rm in}\to64\) channels, producing a tensor of shape \([64\times N\times N]\).
	\item \textbf{Four Fourier layers:} each layer applies a spectral convolution on the lowest 32 modes per dimension with 64 input/output channels and zero-padding (output \([64\times N\times N]\)), followed by a 1×1 convolution (\(64\to64\)) and a GELU activation, defined as \(\mathrm{GELU}(x)=x\,\Phi(x)\) with \(\Phi\) the standard normal cumulative distribution function (CDF) \cite{Hendrycks2016}.
	\item \textbf{Output head:} two successive 1×1 convolutions mapping \(64\to64\to1\), yielding the final output tensor of shape \([1\times N\times N]\).
\end{itemize}

MAD1-FNO1 is trained with two input channels (boundary and mask), whereas MAD1-FNO2 additionally incorporates explicit \(x,y\) coordinate channels.

All models use the Adam optimizer \cite{Kingma2014AdamAM} (learning rate \(10^{-4}\)). Unless explicitly noted in the Results (e.g.\ MAD1-FNO variants for 2000 epochs), all experiments are trained for \(2\times10^5\) epochs.

\textbf{Note:} For the unit disk region, the number of boundary points was set to 100, and the branch network's input layer was adjusted to 100 neurons. For the L-shaped region, the number of boundary points remained 200, and the branch network's input layer was kept at 200 neurons.

\subsection{Evaluation metrics}
The loss function for MAD is defined as the Mean Squared Error (MSE), which quantifies the deviation between predicted and true solution values:
\begin{equation}
	L_{\text{MAD}} = \frac{1}{N M} \sum_{i=1}^{N} \sum_{j=1}^{M} 
	(u_{i,j}^{\text{NN}} - u_{i,j}^{*})^2.
	\label{eq:mad_loss}
\end{equation}
Here, \(N\) and \(M\) represent the number of training functions and spatial sampling points, respectively. \(u_{i,j}^{\text{NN}}\) and \(u_{i,j}^{*}\) denote the neural operator values and true solution values at the \(j\)-th spatial point of the \(i\)-th function. In this study, \(N = 2000\) and \(M = 2601\) are used for the square region, reflecting typical values for sufficient coverage of the solution space. For the unit disk region, \(M = 1957\) is used, and for the L-shaped region, \(M = 1976\).

For PINN-based methods, a weighted loss function is employed to balance the contributions of internal physics-informed and boundary condition terms:
\begin{equation}
	\begin{split}
		L_{\text{PINN}} = \omega_1 \left( \frac{1}{N M_1} \sum_{i=1}^{N} \sum_{j=1}^{M_1} 
		(\mathcal{L}u_{i,j}^{\text{NN}} - f_{i,j})^2 \right) \\ 
		+ \omega_2 \left( \frac{1}{N M_2} \sum_{i=1}^{N} \sum_{j=1}^{M_2} 
		(u_{i,j}^{\text{NN}} - u_{i,j}^{*})^2 \right).
	\end{split}
	\label{eq:pinn_loss}
\end{equation}
Here, \(M_1\) and \(M_2\) represent the spatial sampling points for the residual and boundary conditions, respectively. The operator \(\mathcal{L}\) denotes the differential operator in the PDE, while \(f_{i,j}\) represents the source term. The weights \(\omega_1\) and \(\omega_2\) are used to balance the contributions of the residual and boundary loss terms. In this study, \(N = 2000\), \(M_1 = 2601\), and \(M_2 = 200\) are used for the square region. For the unit disk region, \(M_1 = 1957\) and \(M_2 = 100\), while for the L-shaped region, \(M_1 = 1976\) and \(M_2 = 200\). The weights are set to \(\omega_1 = 0.9\) and \(\omega_2 = 0.1\), balancing the contributions of residual and boundary losses.
	
\section{Results}
\label{sec:results}
This section presents the numerical results that demonstrate MAD’s accuracy
and efficiency on various PDE benchmarks and neural-operator architectures.

In this work, according to the mathematical properties of different types of equations, we propose three MAD variants—MAD0 for equations with source terms, MAD1 (fundamental‐solution based) and MAD2 for source‐free equations—and evaluate them through extensive numerical experiments on the Lapl\-ace, Poisson, and Helmholtz equations to assess their accuracy, efficiency, and generalization ability.

To be concrete, the three equation families under study can be written as follows:
\begin{equation}
	\begin{cases}
		\nabla^{2}u = f, & \text{in } \Omega, \\
		u = g, & \text{on } \partial \Omega,
	\end{cases}
	\label{eq:poisson}
\end{equation}

\begin{equation}
	\begin{cases}
		\nabla^{2}u + k u = f, & \text{in } \Omega, \\
		u = g, & \text{on } \partial \Omega,
	\end{cases}
	\label{eq:helmholtz}
\end{equation}
where \( \Omega = [0, 1]^2 \) denotes the computational domain, \( f \) the source term, \( g \) the boundary condition (the method can be directly applied to any other type of boundary condition), and \( k \) a given positive scalar parameter in the Helmholtz equation, with \( \sqrt{k} \) denoting the wave number, $f$ and $g$ are two input functions that neural operators need to learn.

Leveraging MAD’s rapid, exact analytical data generation—decoupled entirely from network design—our framework can produce large, high‐quality datasets in seconds.  Moreover, it integrates seamlessly with any modern neural architectures or optimization algorithms.  In particular, for each method, we designed two test sets (test set 1 and test set 2) sampled using completely different schemes, thereby providing a more convincing comparison of generalization performance.

To quantify MAD’s dataset–construction efficiency, we generated  
200 training samples for the two-dimensional Helmholtz equation,
\[
\nabla^{2}u + 100\,u = f , \qquad (x,y)\in[0,1]\times[0,1],
\]
on the unit square. For a fair timing comparison, the finite-difference (FD) baseline had to match MAD’s analytic (error-free) accuracy. Using the analytic test function \(u_{\mathrm{ref}}(x,y)=\cos(6x)\sin(8y)\), we obtained the relative \(L^{2}\) errors of a standard five-point central-difference scheme: $\operatorname{err}(h=0.02)=4.10\times10^{-2}$,
$\operatorname{err}(h=0.01)=1.19\times10^{-2}$,
$\operatorname{err}(h=0.002)=5.07\times10^{-4}$, and
$\operatorname{err}(h=0.001)=1.28\times10^{-4}$.

Consequently, the finite‐difference solver was run on the fine grid, \(h = 0.001\). Under this criterion, MAD produced the 200 Helmholtz samples in \textbf{2.19 s} on an Intel Core i7-12800HX CPU, whereas the high-precision FD solver required \textbf{4,663.17 s}, underscoring MAD’s overwhelming speed advantage.

All training and evaluation tasks were run on the ORISE supercomputing system, ensuring adequate computing resources for large-scale modeling.  MAD’s training datasets were generated analytically, providing exact solutions without any numerical approximation error and ensuring diverse coverage;  Supplementary Section S1.1 provides examples of these training functions.

To evaluate MAD's performance, we conducted comparative analyses with PINN-based methods using metrics such as training time, convergence rate, and relative \( L_2 \) error, highlighting the computational advantages of MAD. Training loss curves for different configurations, geometric domains, and problem setups are detailed in Supplementary Section S1.2, demonstrating MAD's superior convergence behavior.

Nearly all data presented in the tables were obtained from the average of four experiments, each consisting of randomly generating two training sets and conducting two independent training sessions on each. The results from repeated training sessions showed minimal variation. Detailed tables, including the range of these variations, are provided in Supplementary Section S2.

	\subsection{General equations with source using MAD0}
	
	In this subsection, we demonstrate the performance of the MAD0 method on the Poisson equation~\eqref{eq:poisson} and the Helmholtz equation~\eqref{eq:helmholtz} with general source terms. Table~\ref{table1} presents the results of solving these equations using MAD0 and PINN-based methods, highlighting the performance in terms of training time, training loss, and relative \( L_2 \) error for both accuracy and computational efficiency. The MAD0 method consistently surpasses the PINN-based method, demonstrating significantly lower training loss and relative \( L_2 \) errors, while achieving a reduction in training time by approximately 30\% to 50\% (with a same amount of epochs).
	
	\begin{table*}[htbp!]
		\centering
		\renewcommand{\arraystretch}{1.5}
		\resizebox{\textwidth}{!}{
			\begin{tabular}{lccccc}
				\hline
				\textbf{Equation} & \textbf{Model} & \parbox[c][1cm]{3.5cm}{\bfseries Training Time\\(s)} & \textbf{Training Loss} & \parbox[c][1.3cm]{4.5cm}{\bfseries Relative L2 Error\\ (Test Set 1)} & \parbox[c][1.3cm]{4.5cm}{\bfseries Relative L2 Error\\(Test Set 2)} \\
				\hline
				$\nabla^{2}u = f, \; u|_{\partial \Omega} = g$ & PINN-based & 1884393.88 & 2.04E-05 & 8.80E-01 & 1.65E-02 \\
				& \textbf{MAD0} & \textbf{1244780.84} & \textbf{2.17E-09} & \textbf{1.12E-02} & \textbf{9.53E-03} \\
				\hline
				$\nabla^{2}u + u = f, \; u|_{\partial \Omega} = g$ & PINN-based & 2479420.71 & 3.90E-06 & 7.92E-01 & 8.59E-03 \\
				& \textbf{MAD0} & \textbf{1430202.14} & \textbf{1.34E-10} & \textbf{4.96E-03} & \textbf{3.72E-03} \\
				\hline
				$\nabla^{2}u + 10u = f, \; u|_{\partial \Omega} = g$ & PINN-based & 2346844.08 & 6.83E-06 & 1.05E+00 & 6.38E-03 \\
				& \textbf{MAD0} & \textbf{1415130.55} & \textbf{1.57E-10} & \textbf{3.53E-03} & \textbf{2.31E-03} \\
				\hline
				$\nabla^{2}u + 100u = f, \; u|_{\partial \Omega} = g$ & PINN-based & 2384881.09 & 1.15E-02 & 6.26E+01 & 9.97E-01 \\
				& \textbf{MAD0} & \textbf{1405285.52} & \textbf{4.97E-09} & \textbf{1.62E-02} & \textbf{7.19E-02} \\
				\hline
			\end{tabular}
		}
		\caption{Comparison of PINN-based methods and MAD0 for solving the Poisson (equation~\eqref{eq:poisson}) and Helmholtz (equation~\eqref{eq:helmholtz}) equations with source. Test Set 1 comprises analytically generated solutions closely aligned with the MAD0 training set, whereas Test Set 2 is created using the five-point central difference method on a \(0.001 \times 0.001\) grid over the \([0,1] \times [0,1]\) domain, ensuring high accuracy for evaluating the performance of different models.}
		\label{table1}
	\end{table*}
	
\begin{figure*}[htbp!]
		\centering
		\includegraphics[width=\textwidth]{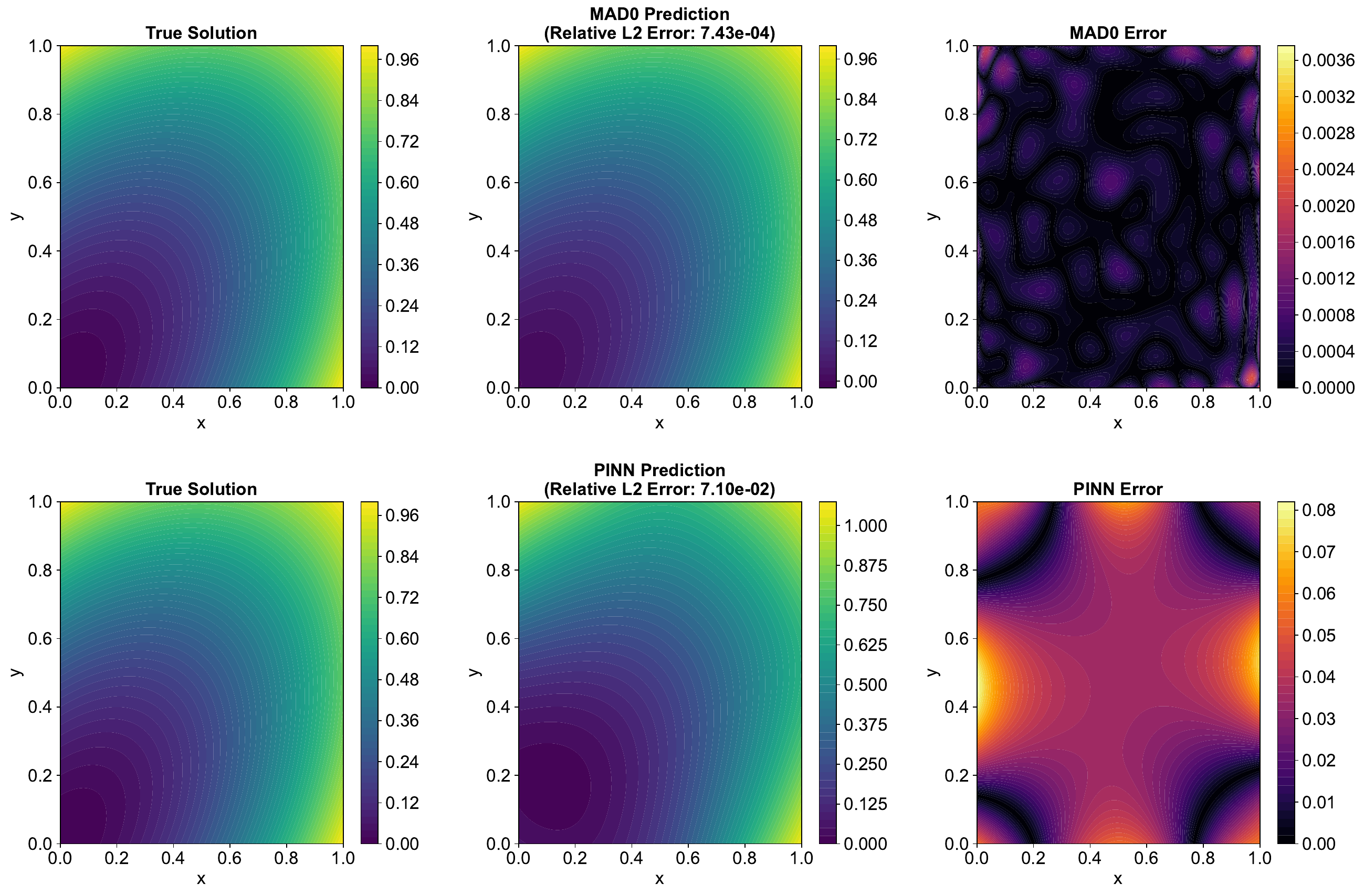}
		\caption{Comparison of the true solution \( u(x, y) = x^2 + y^2 - xy \), MAD0 prediction, PINN-based prediction, and the corresponding error distributions for Example 2 (equation~\eqref{eq:helmholtz}, \( k = 1 \)). Relative \( L_2 \) error is computed as \(\|u_{\text{pred}} - u_{\text{true}}\|_2 / \|u_{\text{true}}\|_2\), based on a grid of \( n = 200 \).}
		\label{fig:example2}
	\end{figure*}

\begin{figure*}[htbp!]
		\centering
		\includegraphics[width=\textwidth]{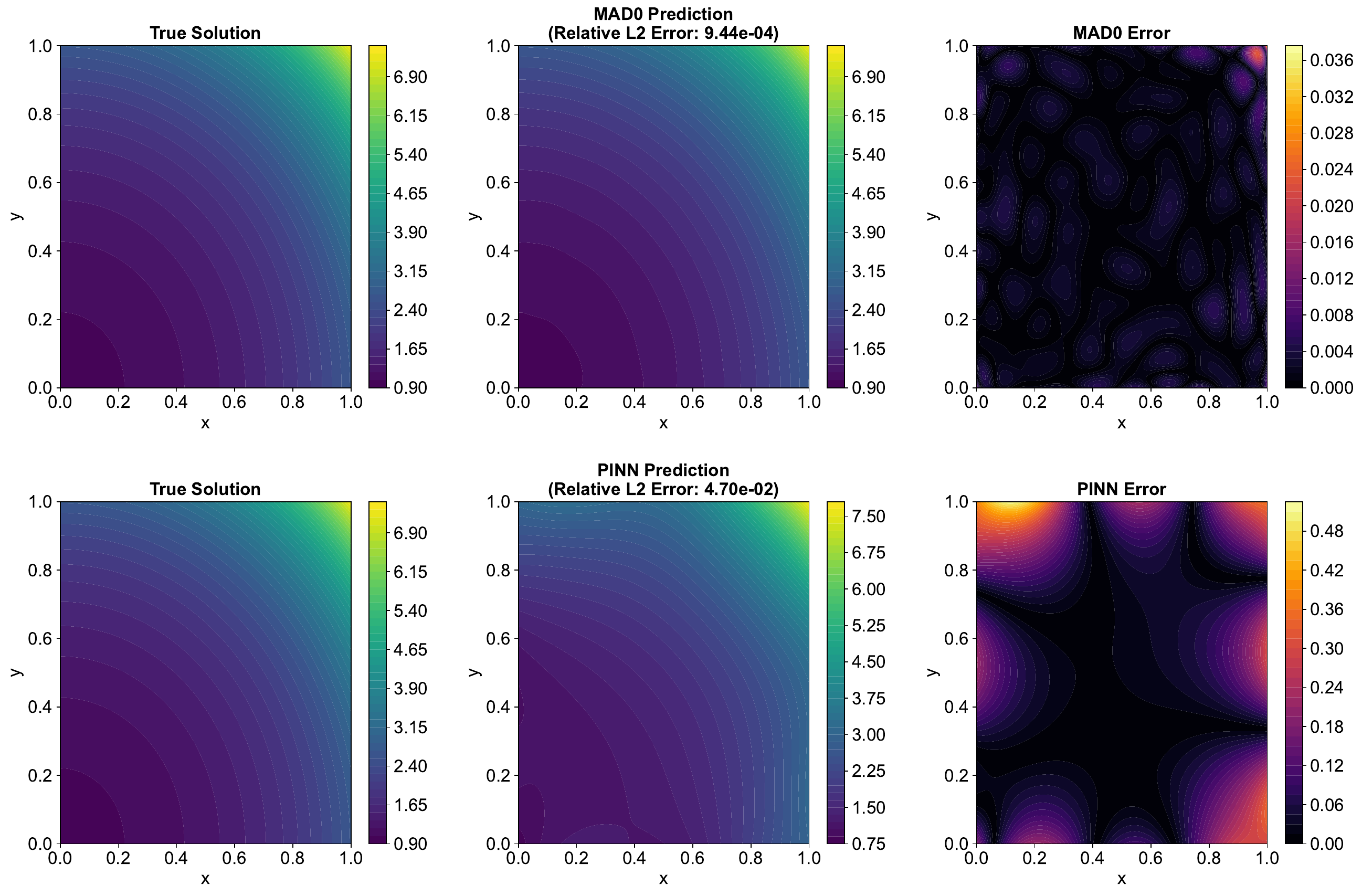}
		\caption{Comparison of the true solution \( u(x, y) = e^{x^2 + y^2} \), MAD0 prediction, PINN-based prediction, and the corresponding error distributions for Example 3 (equation~\eqref{eq:helmholtz}, \( k = 1 \)). Relative \( L_2 \) error is computed as \(\|u_{\text{pred}} - u_{\text{true}}\|_2 / \|u_{\text{true}}\|_2\), based on a grid of \( n = 200 \).}
		\label{fig:example3}
	\end{figure*}

	Figs.~\ref{fig:example2}, and \ref{fig:example3} depict predictions for two representative cases: quadratic and exponential solutions. The MAD0 method exhibits superior accuracy, particularly in regions with steep gradients, underscoring its robustness in handling challenging scenarios. Additional results for other test cases can be found in Supplementary Section S1.3 (Figs. S.10–S.19).
	
	These results underscore the robustness and efficiency of the MAD0 method in solving equations with source. To further explore its potential, we extend the experiments to source-free equations (\( f = 0 \)), where solutions exhibit distinct structural patterns, enabling a broader assessment of MAD's capabilities.
	
\subsection{Source-free equations using MAD1 (based on fundamental solutions)}
For the linear elliptic PDEs such as equations~\eqref{eq:poisson}-\eqref{eq:helmholtz} that have fundamental solutions, the sampled solution data with our MAD1 method exactly satisfy the corresponding source-free equations. In this subsection, we report the operator learning results for Laplace equation and Helmholtz equations with \(f=0\) using MAD1. As shown in Table~\ref{table2}, the MAD1 method consistently outperforms the PINN-based approach across all test configurations, and detailed results for specific cases are provided in Supplementary Section S1.3 (Figs. S.20–S.25). Notably, at \( k = 100 \), MAD1 achieves a relative \( L_2 \) error of \( 1.93 \times 10^{-3} \) on Test Set 2, significantly lower than \( 9.99 \times 10^{-1} \) obtained by PINN-based methods. These results demonstrate the superior capability of MAD1 in capturing oscillatory behaviors typical of the Helmholtz equation without relying on source terms.

\begin{table*}[htbp!]
	\centering
	\renewcommand{\arraystretch}{1.5}
	\resizebox{\textwidth}{!}{
		\begin{tabular}{lccccc}
			\hline
			\textbf{Equation} & \textbf{Model} & \parbox[c][1cm]{3cm}{\bfseries Training Time\\(s)} & \textbf{Training Loss} & \parbox[c][1.3cm]{4.5cm}{\bfseries Relative L2 Error\\(Test Set 1)} & \parbox[c][1.3cm]{4.5cm}{\bfseries Relative L2 Error\\(Test Set 2)} \\
			\hline
			$\nabla^{2}u = 0, \; u|_{\partial \Omega} = g$ & PINN-based & 454660.56 & 1.24E-05 & 3.10E-02 & 1.22E-02 \\
			& \textbf{MAD1} & \textbf{221264.99} & \textbf{1.62E-07} & \textbf{1.77E-03} & \textbf{1.40E-03} \\
			\hline
			$\nabla^{2}u + u = 0, \; u|_{\partial \Omega} = g$ & PINN-based & 500631.87 & 8.75E-06 & 2.59E-02 & 1.04E-02 \\
			& \textbf{MAD1} & \textbf{220906.40} & \textbf{1.31E-07} & \textbf{1.96E-03} & \textbf{1.57E-03} \\
			\hline
			$\nabla^{2}u + 10u = 0, \; u|_{\partial \Omega} = g$ & PINN-based & 498656.61 & 1.27E-05 & 4.22E-02 & 8.87E-03 \\
			& \textbf{MAD1} & \textbf{220841.66} & \textbf{2.65E-07} & \textbf{2.61E-03} & \textbf{1.06E-03} \\
			\hline
			$\nabla^{2}u + 100u = 0, \; u|_{\partial \Omega} = g$ & PINN-based & 517114.62 & 2.11E-02 & 9.74E-01 & 9.99E-01 \\
			& \textbf{MAD1} & \textbf{220475.28} & \textbf{1.22E-06} & \textbf{8.87E-03} & \textbf{1.93E-03} \\
			\hline
		\end{tabular}
	}
	\caption{Comparison of PINN-based and MAD1 methods for solving the equations~\eqref{eq:poisson}-\eqref{eq:helmholtz} with \(f=0\). Each value in the table represents the mean over four independent experiments. Test Set 1 comprises analytically generated solutions closely aligned with the MAD1 training set, whereas Test Set 2 is created using the five-point central difference method on a \(0.001 \times 0.001\) grid over the \([0,1] \times [0,1]\) domain, ensuring high accuracy for evaluating the performance of different models.}
	\label{table2}
\end{table*}

\begin{figure*}[htbp!]
	\centering
	\includegraphics[width=\textwidth]{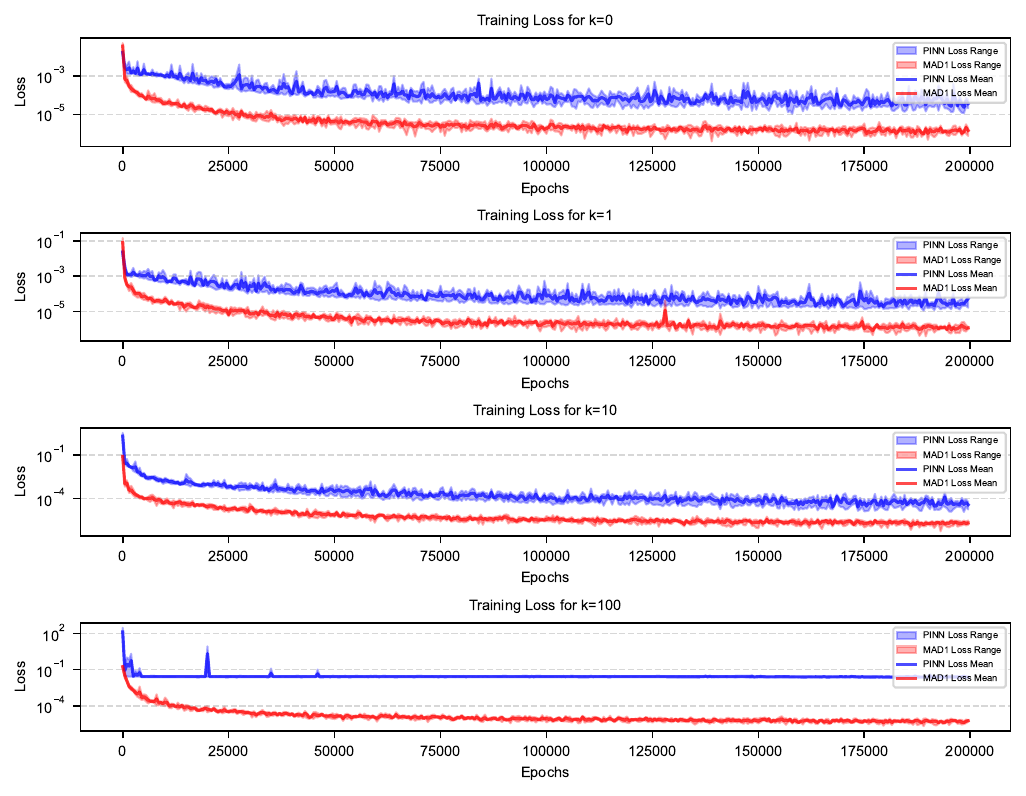}
	\caption{Training loss curves for the Laplace equation (\( k = 0 \)) and the source-free Helmholtz equation at \( k = 1, 10, 100 \). Each panel shows the loss for both PINN and MAD1 methods during training, with the \( y \)-axis plotted in log-scale to emphasize convergence rates. Subfigures correspond to different \( k \) values: \textbf{Top}: Laplace equation (\( k = 0 \)), \textbf{Upper Middle}: \( k = 1 \), \textbf{Lower Middle}: \( k = 10 \), \textbf{Bottom}: \( k = 100 \).}
	\label{fig:loss_helmholtz}
\end{figure*}

The loss curve shown in Fig.~\ref{fig:loss_helmholtz} represents the results of four independent training runs for clarity.

Fig.~\ref{fig:loss_helmholtz} illustrates the convergence behavior for different \( k \) values, including the Laplace equation (i.e., the source-free Helmholtz equation with \( k = 0 \)). The training loss curves show that MAD1 not only converges faster but also achieves significantly lower final losses, highlighting its computational efficiency. As \( k \) increases, the PINN-based method faces difficulties in maintaining convergence, whereas MAD1 consistently adapts to the increased complexity. Notably, for the Laplace equation (\( k = 0 \)), MAD1 demonstrates superior performance compared to the PINN-based approach, further validating its robustness across different problem settings.

These observations suggest that the MAD1 method is particularly well-suited for solving equations where the solution relies heavily on boundary conditions, making it a promising alternative for applications requiring high precision in similar scenarios. Notably, for \( k=100 \), the training loss of the PINN-based method stagnates early, indicating significant challenges in capturing the highly oscillatory solutions of the Helmholtz equation at larger \(k\) values. This limitation is likely attributed to the loss landscape of PINN, where gradients for such problems become highly irregular or vanish, leading to suboptimal optimization performance. In contrast, the MAD1 method leverages its supervised data-driven framework and high-quality training datasets generated from fundamental solutions, facilitating smoother optimization and effective learning of the solution.

The observed efficiency and adaptability of the MAD1 method lay a strong foundation for exploring its performance across different geometric domains, which will be compared with MAD2 in the next subsection.

\subsection{Geometric adaptability: solving Laplace equations using MAD1 and MAD2}
To evaluate the adaptability of MAD across various geometric domains, we tested its performance on the Laplace equation in three representative configurations: square, unit disk, and L-shaped regions. Additionally, we propose another artificial data generation method for the Laplace equation, named MAD2, as a comparison to MAD1. MAD2 employs trigonometric and hyperbolic functions to generate harmonic functions satisfying the Laplace equation, providing an alternative approach to data generation.

As shown in Table~\ref{table4}, MAD1 consistently demonstrates strong performance across various test sets and geometric domains. In contrast, MAD2 achieves higher accuracy on Test Set 2, especially in the square domain, but its performance on Test Set 1, although still superior to PINN-based methods, shows higher relative errors. These results highlight MAD1’s superior generalization ability across different test sets and geometries.

\begin{table*}[htbp!]
	\centering
	\renewcommand{\arraystretch}{1.5}
	\resizebox{\textwidth}{!}{
	\begin{tabular}{lccccc}
		\hline
		\textbf{Geometry} & \textbf{Model} & \parbox[c][1cm]{3cm}{\bfseries Training Time\\(s)} & \textbf{Training Loss} & \parbox[c][1.3cm]{4.5cm}{\bfseries Relative L2 Error \\ (Test Set 1)} & \parbox[c][1.3cm]{4.5cm}{\bfseries Relative L2 Error \\ (Test Set 2)} \\
		\hline
		\multirow{3}{*}{\includegraphics[width=1.5cm]{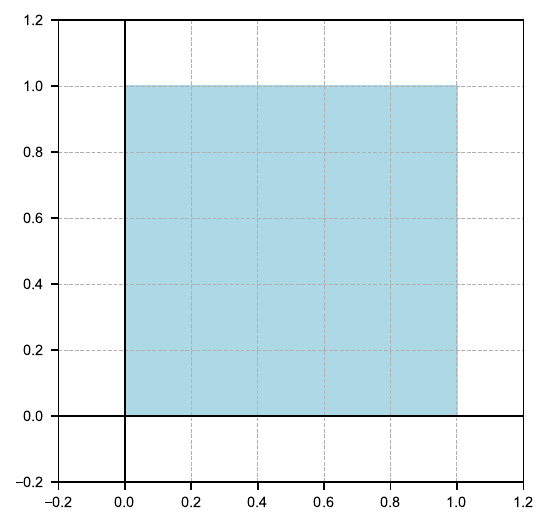}} & PINN-based & 454660.56 & 1.24E-05 & 3.10E-02 & 2.14E-02 \\
		& \textbf{MAD1} & \textbf{221264.99} & \textbf{1.62E-07} & \textbf{1.77E-03} & \textbf{4.37E-04} \\
		& \textbf{MAD2} & \textbf{221680.32} & \textbf{4.75E-09} & \textbf{1.74E-02} & \textbf{1.57E-04} \\
		\hline
		\multirow{3}{*}{\includegraphics[width=1.5cm]{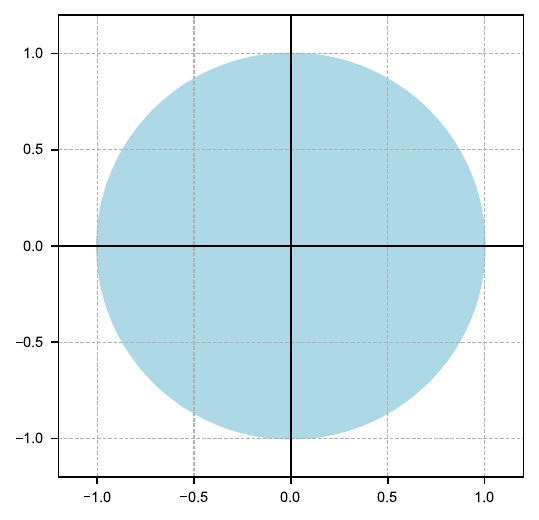}} & PINN-based & 311953.14 & 2.08E-06 & 4.10E-02 & 2.99E-02 \\
		& \textbf{MAD1} & \textbf{149158.02} & \textbf{5.89E-07} & \textbf{4.06E-03} & \textbf{1.33E-03} \\
		& \textbf{MAD2} & \textbf{149013.98} & \textbf{8.42E-09} & \textbf{1.95E-02} & \textbf{2.78E-04} \\
		\hline
		\multirow{3}{*}{\includegraphics[width=1.5cm]{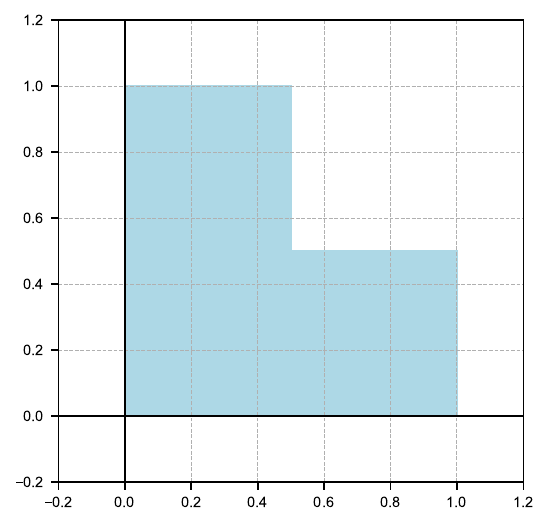}} & PINN-based & 358509.82 & 1.76E-03 & 2.32E-01 & 2.05E-01 \\
		& \textbf{MAD1} & \textbf{176299.15} & \textbf{1.55E-07} & \textbf{2.08E-03} & \textbf{4.00E-04} \\
		& \textbf{MAD2} & \textbf{174649.07} & \textbf{4.92E-09} & \textbf{4.12E-02} & \textbf{1.68E-04} \\
		\hline
	\end{tabular}
    }
	\caption{Performance of PINN-based and MAD methods on the Laplace equation (equation~\eqref{eq:poisson}, $f=0$) for different geometric domains. MAD1 and MAD2 refer to models trained on fundamental solution datasets and trigonometric/hyperbolic function datasets, respectively. Test Set 1 corresponds to MAD1 training data, and Test Set 2 corresponds to MAD2 training data.}
	\label{table4}
\end{table*}

The performance difference between MAD1 and MAD2 can be attributed to their distinct data generation methods. MAD1 relies on fundamental solutions, which are inherently tied to the Laplace equation and provide uniform coverage across various geometric domains. This approach ensures that the generated solutions are physically meaningful and generalizable. On the other hand, MAD2 uses linear combinations of trigonometric and hyperbolic functions to construct harmonic solutions. While this method is expressive in specific test sets like Test Set 2, it lacks the comprehensive coverage of the harmonic function space, leading to reduced generalization performance.

Moreover, while MAD-based methods demonstrate strong adaptability across different geometric domains, PINN-based methods exhibit notable limitations. As evidenced in Table~\ref{table4}, PINN-based methods maintain relative \( L_2 \) errors around the \( 10^{-2} \) level in simpler domains like the square and unit disk but show significant performance degradation in the L-shaped region, with relative errors increasing to \( 2.32 \times 10^{-1} \). This discrepancy is partly due to the fact that the PINN method used here applies the same basic sampling strategy across all test cases. While PINN-based methods can also improve accuracy by adjusting the sampling strategy—such as employing non-uniform or adaptive point distributions—our comparisons focus on results obtained using the most straightforward and uniform sampling approach to ensure a fair evaluation. In contrast, MAD methods benefit from their structured, physically-informed datasets, enabling superior performance and generalization across diverse domains.

These findings not only underscore the advantages of MAD1 in generalization and accuracy but also highlight the critical role of data generation methods in enhancing geometric adaptability. In the next subsection, we explore how variations in neural network configurations, such as depth and width, further influence the performance of MAD methods.

\subsection{Impact of neural network configurations}
To investigate the impact of neural network architectures on performance, we evaluated MAD and PINN-based methods on the Laplace equation (equation~\eqref{eq:poisson} with \(f = 0\)).

It is important to note that both test sets were not generated using the five-point central difference method, which had been discussed in previous subsections. Instead, the evaluation focused on the performance of MAD and PINN-based methods across different neural network (NN) configurations.

The wide configuration increased the number of neurons per layer by doubling each hidden layer’s width compared to the baseline. The deep configuration added more layers to the trunk network by doubling the number of hidden layers compared to the baseline (detailed network structure see Section~\hyperref[sec:methods]{4}). As summarized in Table~\ref{table3}, MAD consistently outperformed PINN-based methods in both accuracy and training efficiency across all configurations. Specifically, MAD1 demonstrated balanced performance across both test sets, while MAD2 showed stronger accuracy on Test Set 2 but less competitive performance on Test Set 1, indicating a trade-off between accuracy and generalization capability.

\begin{table*}[htbp!]
	\centering
	\renewcommand{\arraystretch}{1.5}
	\resizebox{\textwidth}{!}{
		\begin{tabular}{lcccccc}
			\hline
			\textbf{NN Configuration} & \textbf{Model} & \parbox[c][1cm]{3cm}{\bfseries Training Time\\ (s)} & \textbf{Training Loss} & \parbox[c][1.3cm]{4.5cm}{\bfseries Relative L2 Error\\ (Test Set 1)} & \parbox[c][1.3cm]{4.5cm}{\bfseries Relative L2 Error\\(Test Set 2)} \\
			\hline
			Baseline NN& PINN-based & 454660.56 & 1.24E-05 & 3.10E-02 & 2.14E-02  \\
			& \textbf{MAD1} & \textbf{221264.99} & \textbf{1.62E-07} & \textbf{1.77E-03} & \textbf{4.37E-04} \\
			& \textbf{MAD2} & \textbf{221680.32} & \textbf{4.75E-09} & \textbf{1.74E-02} & \textbf{1.57E-04} \\
			\hline
			Wide NN& PINN-based & 972619.33 & 1.05E-05 & 3.19E-02 & 2.20E-02 \\
			& \textbf{MAD1} & \textbf{330268.64} & \textbf{1.31E-07} & \textbf{1.53E-03} & \textbf{3.85E-04} \\
			& \textbf{MAD2} & \textbf{328506.57} & \textbf{6.22E-09} & \textbf{1.80E-02} & \textbf{1.85E-04} \\
			\hline
			Deep NN& PINN-based & 696686.96 & 2.68E-05 & 3.00E-02 & 1.99E-02 \\
			& \textbf{MAD1} & \textbf{235727.07} & \textbf{4.54E-08} & \textbf{1.12E-03} & \textbf{2.27E-04} \\
			& \textbf{MAD2} & \textbf{235027.55} & \textbf{1.92E-09} & \textbf{1.73E-02} & \textbf{1.00E-04} \\
			\hline
		\end{tabular}
	}
	\caption{Comparison of PINN-based methods and MAD across different neural network configurations for solving the Laplace equation (equation~\eqref{eq:poisson}, \(f=0\)). MAD1 and MAD2 refer to models trained on fundamental solution datasets and trigonometric/hyperbolic function datasets, respectively. Test Set 1 corresponds to MAD1 training data type, and Test Set 2 aligns with MAD2 data type.}
	\label{table3}
\end{table*}

While PINN-based methods maintained relative \( L_2 \) errors around the \( 10^{-2} \) level across different configurations, their computational cost increased significantly with the complexity of the network architectures. In contrast, MAD consistently achieved superior accuracy and efficiency without a corresponding increase in computational cost. For instance, MAD1 reduced training time by 66.2\% in the deep configuration compared to PINN-based methods. These results underscore MAD's superior computational efficiency and its ability to effectively scale to larger network architectures while maintaining high performance.

These findings emphasize MAD’s strong adaptability across different neural network configurations and demonstrate its efficiency in solving large-scale computational problems. 

\subsection{Experiments using the FNO network architecture}

As a demonstration outside of the DeepOnet framework, we also trained a standard Fourier neural operator (FNO) using MAD synthesis data of the 2D Laplace equation on \([0,1]^2\). Traditionally, such data are obtained by running high-fidelity numerical solvers on fine meshes, which incurs significant computational cost. In contrast, our MAD framework generates exact analytic solutions almost instantly.

MAD1-FNO1 was trained using two input channels (boundary and mask), whereas MAD1-FNO2 additionally incorporated explicit \(x,y\) coordinate channels into the input.

\begin{table*}[htbp]
	\centering
	\small
	\renewcommand{\arraystretch}{1.3}
	\setlength{\tabcolsep}{12pt}
	\resizebox{0.8\textwidth}{!}{
		\begin{tabular}{lccc}
			\hline
			\textbf{Variant}                    & \textbf{Epochs} & \textbf{Training Loss} & \textbf{Relative \(L_2\) Error} \\
			\hline
			MAD1-FNO1                           & 2\,000          & \(1.59\times10^{-6}\)  & \(2.82\times10^{-3}\)           \\
			MAD1-FNO2                           & 2\,000          & \(1.56\times10^{-6}\)  & \(2.81\times10^{-3}\)           \\
			\hline
		\end{tabular}
	}
	\caption{Training details and final errors of the two MAD1-FNO variants on the 2D Laplace equation.}
	\label{tab:fno_extension}
\end{table*}

As shown in Table~\ref{tab:fno_extension}, both variants converge to sub-0.3\% relative \(L_2\) error within only 2\,000 epochs, demonstrating that MAD-generated data can be seamlessly integrated into the FNO architecture.

\subsection{A test of solving the 3D Laplace equation}

We also evaluated the MAD1 framework combined with a DeepONet backbone on the 3D Laplace equation over \([0,1]^3\). After 10\,000 training epochs, the model achieved a training loss of \(1.61\times10^{-5}\) and a relative \(L_2\) error of \(2.03\times10^{-2}\). These results demonstrate that our MAD framework can be readily extended to higher-dimensional boundary-value problems.

This example is just showing a three-dimensional case. General three-dimens\-ional problems require much greater computational costs and/or further methodological research, which is a future research interest and focus.

\section{Discussion}
﻿\label{sec:discussion}
Our results demonstrate the effectiveness of MAD across various experimental settings, including different PDEs, neural network architectures, and geometric domains. Compared to PINN-based methods, MAD achieves faster convergence during training and lower relative \(L_2\) errors, while maintaining strong generalization. MAD1, using a dataset based on fundamental solutions, shows strong adaptability to complex geometric shapes (e.g., L-shaped domains) and diverse network architectures, highlighting MAD's robustness and flexibility in handling geometric complexity and integrating with different neural network designs.

MAD introduces a new paradigm for solving differential equations by generating analytically derived datasets that are either error-free (e.g., exact solutions like Laplace fundamental solutions) or near-error-free with controllable precision (e.g., truncated series or numerical integrals). This approach circumvents the high costs and uncertainties associated with numerical simulations, providing high-fidelity data samples that alleviate key computational bottlenecks in operator learning. Unlike conventional Neural Operator methods, which rely on extensive precomputed datasets, MAD achieves both efficiency and accuracy through its noise-free approach, particularly when scaling to larger domains or complex operators.

MAD’s derivative-free nature makes it well-suited for integration with advanced neural network architectures, such as transformer-based models \cite{10.5555/3295222.3295349}, Fourier Neural Operators (FNO) \cite{li2021fourier}, and hybrid approaches combining physics-informed and data-driven paradigms. Transformer models excel at capturing long-range dependencies, while FNO leverages spectral convolutions to efficiently solve PDEs with complex domains and multi-scale features. Combining MAD’s high-fidelity datasets with these architectures improves training efficiency and accuracy. In our experiments, the MAD1-FNO2 model, trained to solve the Laplace equation, achieves a relative error of \(2.81 \times 10^{-3}\) after only 2\,000 epochs, demonstrating the convenience of MAD in adapting to various network architectures. Additionally, techniques such as adaptive sampling \cite{TANG2023111868}, importance sampling \cite{ZHANG2025113561}, and multi-scale feature learning \cite{Suo2024} can further enhance MAD's efficiency and precision.

MAD can also integrate naturally with physical data or real-world observations, creating a hybrid framework that combines the strengths of both data-driven and model-driven approaches. This synergy offers exciting opportunities for extending MAD to scenarios where operator approximation is critical or real-world constraints must be incorporated.

The success of MAD, a new physics-embedded data-driven paradigm, relies on the representativeness of the sampled function space and its coverage of the true solution space. Theoretical guarantees support the completeness of neural networks and fundamental solutions, confirming the feasibility of MAD (both MAD0 and MAD1). However, the sampling strategy remains crucial and may significantly impact efficiency and accuracy. Future work in this direction should evaluate the trade-off between sample size, computation time, and accuracy, and explore other effective MAD methods and sampling techniques, as well as the network architectures and optimization methods.

\section*{CRediT authorship contribution statement}
B.Z.L.\ conceived and supervised the project. H.W.\ implemented the code and performed the computations. Both authors analysed the results and wrote the manuscript.

\section*{Declaration of competing interest}
The authors declare that they have no competing interests.

\section*{Acknowledgements}
This research was funded by the Strategic Priority Research Program of the Chinese Academy of Sciences (Grant No.\ XDB0500000) and the National Natural Science Foundation of China (Grant No.\ 12371413).

The numerical calculations were performed on the ORISE Supercomputer.


We thank Sheng Gui for valuable suggestions on graphic design and text editing.



	
\clearpage
\appendix
\addcontentsline{toc}{section}{Supplementary Material}
\includepdf[pages=-,fitpaper=true]{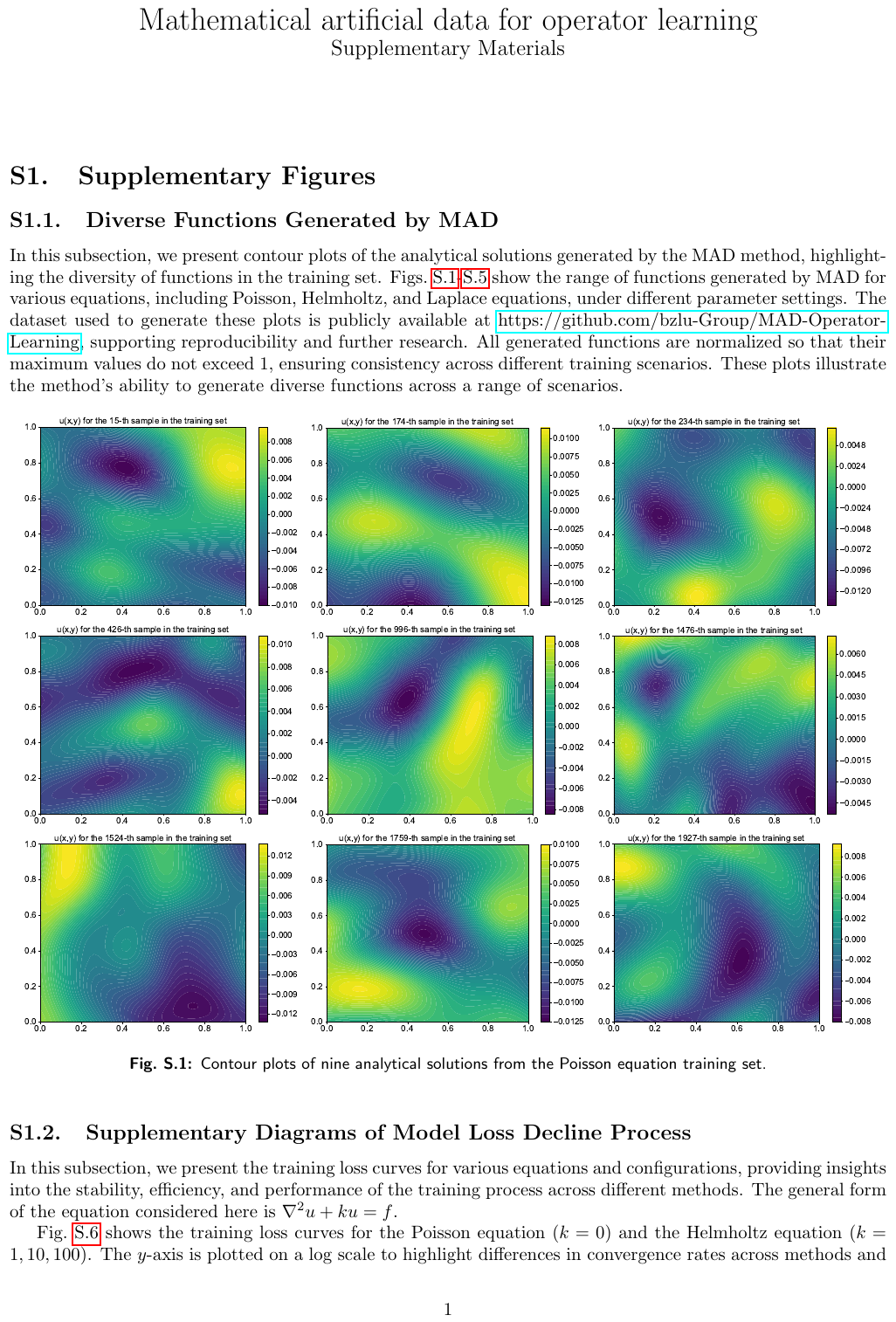}
\end{document}